\documentclass{article}

\usepackage{arxiv}

\usepackage[utf8]{inputenc} 
\usepackage[T1]{fontenc}    
\usepackage{hyperref}       
\usepackage{url}            
\usepackage{booktabs}       
\usepackage{amsfonts}       
\usepackage{nicefrac}       
\usepackage{microtype}      
\usepackage{lipsum}
\usepackage{graphicx}
\graphicspath{ {./images/} }
\usepackage{subcaption} 
\usepackage{amsmath,amssymb,amsfonts}
\usepackage{cleveref}
\usepackage{multirow}
\usepackage{eso-pic}

\title{Geometric Self-Supervised Pre-training for Neural Combinatorial Optimization}

\author{
    David Aguado,
    Daniel Fuertes,
    Carlos R. del-Blanco,
    Fernando Jaureguizar \\
    Grupo de Tratamiento de Imágenes,
    Information Processing and Telecomunications Center,\\
    ETSI Telecomunicación, Universidad Politécnica de Madrid, 28040, Madrid, Spain\\
    \texttt{david.aguado@alumnos.upm.es, \{d.fcoiras, carlosrob.delblanco, fernando.jaureguizar\}@upm.es} \\
}

\begin{document}
\AddToShipoutPictureBG*{
  \AtPageUpperLeft{
    \put(0,-40){
      \makebox[\paperwidth][c]{
        \small This work has been submitted to the IEEE for possible publication. Copyright may be transferred without notice, after which this version may no longer be accessible.
      }
    }
  }
}
\maketitle
\begin{abstract}
Neural Combinatorial Optimization (NCO) techniques have emerged as a highly efficient alternative to traditional exact algorithms for solving routing problems such as the Traveling Salesman Problem (TSP). However, the generalization capabilities of these Reinforcement Learning-based models are severely hindered when scaling to high-dimensional instances. This issue has been mitigated in other domains, like computer vision and natural language processing, by adopting a self-supervised pre-training strategy. Nevertheless, its application to routing graphs, which lack complex topological attributes beyond 2D spatial coordinates, remains a challenge. In this paper, we propose a geometric self-supervised pre-training framework specifically designed to capture spatial invariance and global relative distance distributions. By applying isometric transformations, such as rotations and axial reflections, the model learns robust structural representations prior to the policy optimization phase. Empirical results demonstrate that this strategy consistently outperforms models trained from scratch (baselines), achieving a 7.23\% improvement in tour length for massive zero-shot extrapolation scenarios (TSP1,000). Furthermore, the proposed model exhibits remarkable computational efficiency, delivering speedups of up to two orders of magnitude over the exact solver Concorde at massive scales.

The source code and pre-trained models are publicly available at \url{https://github.com/davidaguadocosano/TSP-GeoPretrain.git}.
\end{abstract}


\section{Introduction} \label{sec:introduction}
Recently, Neural Combinatorial Optimization (NCO) has emerged as a disruptive paradigm to tackle classic NP-hard routing tasks, such as the Traveling Salesman Problem (TSP). By leveraging Deep Reinforcement Learning (DRL), NCO models learn resolution strategies directly from data without requiring labels \cite{DRLNCO}. Once trained, these neural solvers provide near-instantaneous inference, offering a highly efficient alternative for real-time routing applications.

Despite their success on small to medium-scale instances, current NCO models face a severe generalization challenge \cite{kool2018attention, RRNCO}. When trained from scratch, neural agents struggle to learn invariant structural representations, often overfitting to the specific graph scale seen during training. Consequently, their routing performance degrades drastically when exposed to zero-shot extrapolation scenarios, for example, training on 50 nodes and testing on massive graphs of up to 1,000 nodes.

In other domains, such as Natural Language Processing (NLP) \cite{bert} and Computer Vision \cite{mae}, self-supervised pre-training has become the standard mechanism to build robust, generalizable models before fine-tuning on specific tasks. While incorporating these techniques into NCO could theoretically mitigate the generalization bottleneck, specific pre-training frameworks for routing problems are virtually non-existent in the current literature. The reason is that existing graph pre-training methods \cite{mae2, dgi, graphcl} are primarily designed for attribute-rich networks and rely on predicting missing semantic features. This renders them fundamentally incompatible with TSP instances, which are fully connected, attribute-poor graphs where the only meaningful features are 2D spatial coordinates.

To bridge this critical gap, this paper introduces a novel geometric self-supervised pre-training framework specifically tailored for Neural Combinatorial Optimization in routing problems. Based on the premise that the optimal sequence of a TSP remains invariant under isometric transformations that preserve relative distances, we force the network to learn robust spatial symmetries prior to the Reinforcement Learning (RL) policy optimization. To rigorously evaluate this framework, we conduct an extensive ablation study isolating the effects of spatial translations, rotations, and axial reflections. The empirical results demonstrate that our proposed geometric pre-training significantly enhances zero-shot extrapolation capabilities on large-scale instances, maintaining near-instantaneous inference times and establishing a highly scalable neural solver for massive routing networks.

\section{Related Work} \label{sec:related_work}
This section reviews the fundamental literature that contextualizes our proposed approach. First, we outline the evolution of routing problem resolution, transitioning from traditional exact solvers and heuristics to modern Neural Combinatorial Optimization (NCO) architectures. Subsequently, we examine the current landscape of self-supervised pre-training on graphs, highlighting the structural limitations that prevent existing frameworks from being directly applied to attribute-poor topologies such as the TSP.

\subsection{Neural Combinatorial Optimization for Routing}
Traditionally, Vehicle Routing Problems (VRPs) have been addressed using exact mathematical solvers, such as Concorde \cite{concorde}, which rely on branch-and-cut algorithms \cite{branchandbound}. While providing optimal guarantees, their exponential time complexity makes them unsuitable for large-scale or real-time applications. To mitigate this, expert-designed heuristics like the Lin-Kernighan algorithm (LKH-3) \cite{LKH-3} or 2-opt \cite{2op} were developed, trading optimality for computational speed.

In recent years, Neural Combinatorial Optimization (NCO) has emerged as an alternative, leveraging deep learning to parameterize routing policies \cite{NCOreview}. Early attempts utilized Sequence-to-Sequence (Seq2Seq) models and Pointer Networks \cite{pointer} to output valid permutations of input cities. However, treating the TSP as a sequential permutation task introduces unnatural inductive biases, as graphs do not have an inherent sequential order.

To properly address topological relationships, the field transitioned towards permutation-invariant architectures, primarily Graph Neural Networks (GNNs) and Transformer-based models. Approaches using Graph Convolutional Networks (GCN) \cite{gcn}, Graph Attention Networks (GAT) \cite{gat}, and Gated Graph Neural Networks \cite{GatedGCN} allowed the encoders to explicitly process the structural connectivity of the cities. These representations are typically coupled with attention-based decoders and optimized via DRL algorithms. For instance, the Attention Model proposed by Kool et al. \cite{kool2018attention} demonstrated how these architectures can process global structural contexts to construct routing solutions autoregressively. 

Despite these architectural advances, the efficiency of these DRL models remains heavily dependent on the quality of the latent representations extracted from scratch, often leading to performance degradation when scaling to larger graph sizes \cite{generalization}.

\subsection{Self-Supervised Pre-training on Graphs}
Self-supervised pre-training has been extended to Graph Learning to overcome the scarcity of labeled data and improve model generalization. More specifically, state-of-the-art graph pre-training strategies typically fall into generative or contrastive approaches. 

Generative methods, such as GraphMAE \cite{maee} and its subsequent iteration GraphMAE2 \cite{mae2}, apply masked auto-encoding to predict hidden node attributes or edge connectivity. In contrast, approaches such as Deep Graph Infomax (DGI) \cite{dgi} and GraphCL \cite{graphcl} maximize agreement between augmented views to extract transferable representations. Beyond these canonical formulations, transformation-based pretext tasks learn equivariant representations by reconstructing controlled changes in graph topology \cite{gao2023topology}. More recent graph contrastive learning methods emphasize that view generation must be adapted to the input graph and preserve task-relevant information. In particular, GPA learns graph-specific augmentation policies \cite{zhang2024personalized}, AEGCL combines contrastive learning with graph autoencoding in both the topology and attribute domains \cite{li2024aegcl}, and multi-scale injective augmentation enriches node representations by integrating subgraph-, graph-, and node-level contextual information \cite{zhang2024multiscale}. Invariant graph learning has also been investigated for out-of-distribution generalization; for example, GI-Graph uses a diffusion model to generate diverse environment subgraphs for invariant representation learning \cite{zhang2025gigraph}. Collectively, these works show that the design of the pretext task and augmentation mechanism is central to the quality, robustness, and transferability of the learned graph representations.

However, a fundamental limitation exists when applying these standard frameworks to pure routing problems like the TSP. Existing methods are primarily designed for attribute-rich networks (e.g., molecular graphs with atom types, or social networks with user features) and sparse topologies \cite{augmentation}. Conversely, a TSP instance is an attribute-poor, fully connected graph where the only meaningful information is the 2D geometric coordinates of the nodes. Masking attributes or dropping random edges destroys the underlying metric space required to solve the problem. Consequently, these standard techniques are intrinsically incompatible with NCO, presenting a need for geometric-specific pre-training strategies that exploit spatial symmetries and relative distances.

\section{Proposed Framework} \label{sec:methodology}
This section presents the formal definition of the routing task and details the proposed neural architecture. Furthermore, we comprehensively formulate our geometric self-supervised pre-training strategy and the subsequent RL optimization process.

\subsection{Problem Formulation}
The Traveling Salesman Problem (TSP) is mathematically modeled on a complete graph $\mathcal{G} = (\mathcal{V}, \mathcal{E})$, where $\mathcal{V} = \{1, \dots, n\}$ denotes the set of $n$ nodes (or cities) and $\mathcal{E} = \{e_{ij} \ | \ i,j \in \mathcal{V}, i \neq j\}$ represents the edges connecting them. In the standard two-dimensional Euclidean TSP, each node $i$ is characterized exclusively by its spatial coordinates $x_i \in [0, 1]^2$. Each edge $e_{ij}$ has an associated cost $d_{ij}$ corresponding to the Euclidean distance between nodes $i$ and $j$, such that $d_{ij} = \| x_i - x_j \|_2$.

To formalize the optimization objective, let $x_{ij} \in \{0, 1\}$ be a binary decision variable that equals $1$ if the agent travels directly from node $i$ to node $j$, and $0$ otherwise. The goal is to minimize the total length of the tour, expressed by the objective function:
\begin{equation}
    \min \sum_{i=1}^{n} \sum_{j \neq i} d_{ij} x_{ij}
\end{equation}

To ensure that the sequence forms a single continuous, Hamiltonian cycle that visits every node exactly once, the problem is subject to the following constraints:
\begin{equation}
    \sum_{i=1, i \neq j}^{n} x_{ij} = 1 \quad \forall j \in \mathcal{V}
\end{equation}
\begin{equation}
    \sum_{j=1, j \neq i}^{n} x_{ij} = 1 \quad \forall i \in \mathcal{V}
\end{equation}

Additionally, to prevent the formation of disjoint sub-tours, we incorporate the Miller-Tucker-Zemlin (MTZ) formulation \cite{ExplainTSP}. This approach introduces auxiliary continuous variables $u_i \in \mathbb{R}$ for $i=2, \dots, n$, which represent the visitation order of the nodes:
\begin{equation}
    u_i - u_j + n x_{ij} \le n - 1 \quad \forall i, j \in \{2, \dots, n\}, i \neq j
\end{equation}

From a machine learning perspective, the neural solver aims to construct a valid permutation $a = (a_1, a_2, \dots, a_n)$ of the input nodes that satisfies all the aforementioned mathematical constraints while minimizing the sequence length:
\begin{equation}
    L(a) = \sum_{t=1}^{n-1} \| x_{a_t} - x_{a_{t+1}} \|_2 + \| x_{a_n} - x_{a_1} \|_2
\end{equation}
This formal representation serves as the foundational geometric environment upon which our neural architecture and self-supervised pre-training framework are built.

\subsection{Neural Architecture for Routing}
\label{subsec:architecture}

\begin{figure*}[!t]
\centering
\includegraphics[scale=0.9]{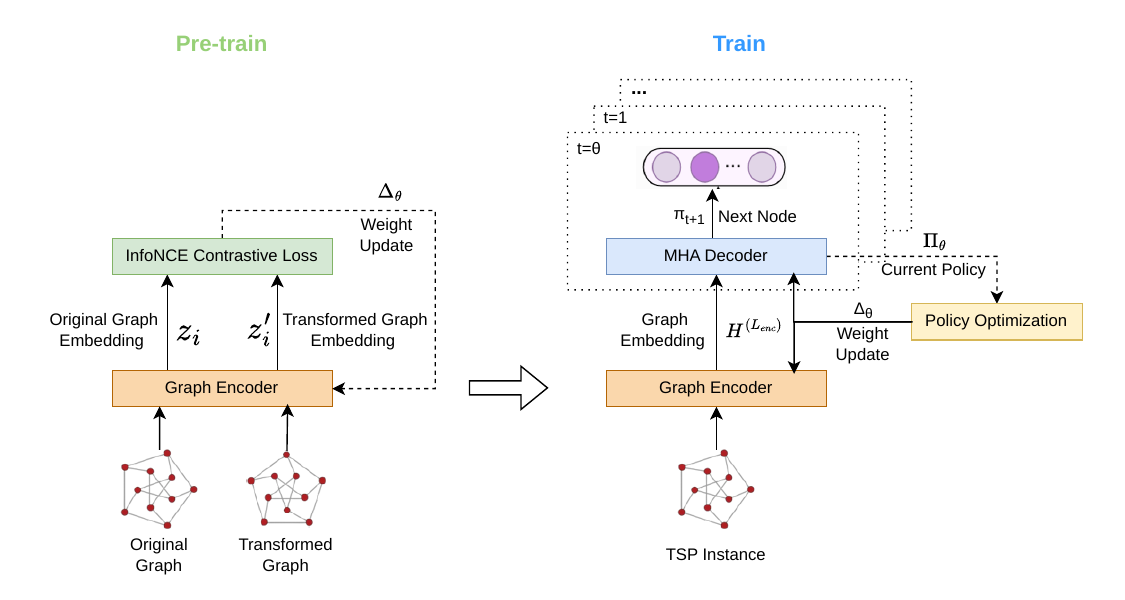}
\caption{Overview of the proposed neural architecture. The system comprises a GatedGCN encoder for structural feature extraction and a Multi-Head Attention decoder for autoregressive route construction, linked by the geometric self-supervised pre-training module.}
\label{fig:architecture}
\end{figure*}

To map the continuous geometric space into a discrete routing policy, we employ an encoder-decoder architecture driven by DRL. As depicted in Figure \ref{fig:architecture}, the system processes the raw 2D coordinates through an anisotropic graph encoder to extract robust topological features, which are subsequently utilized by an autoregressive attention decoder to construct the Hamiltonian cycle step by step. Next, both encoder and decoder are explained in detail.

\subsubsection{Graph Encoder}
The encoding process transforms the TSP structure into a high-dimensional latent space. A critical requirement for routing problems is the ability to process not only the node features but also the explicit edge weights (distances). To achieve this, we utilize a Gated Graph Convolutional Network (GatedGCN) \cite{GatedGCN}, which captures the graph structure through anisotropic message-passing mechanisms, weighting the relevance of connections via edge gating.

Since the TSP is natively defined on a fully connected graph, processing the entire dense edge matrix is computationally prohibitive at scale. To mitigate this computational and memory bottleneck, the graph is sparsified by restricting the connections strictly to the $k$-nearest neighbors ($k$-NN) for each node. The encoding process then begins by linearly projecting the node coordinates $x_i \in [0, 1]^2$ and the Euclidean distances $d_{ij}$ of this sparsified $k$-NN graph into a $d$-dimensional space. Formally, the hidden state of a node $h_i^l$ and the latent edge representation $\hat{e}_{ij}^l$ at layer $l+1$ are updated following an anisotropic network scheme with residual connections. First, we compute the aggregated message $m_i^l$ from the neighborhood:
\begin{equation}
    m_i^l = \text{MAX}_{j \in \mathcal{N}_i}\left(\sigma(\hat{e}_{ij}^l) \odot V^l h_j^l\right),
    \label{eq:message_aggregation}
\end{equation}
where $\mathcal{N}_i$ is the set of neighbors of node $i$ defined by the $k$-NN graph, $\sigma$ represents the sigmoid function acting as a gating mechanism, $V^l \in \mathbb{R}^{d \times d}$ is a learnable weight matrix of layer $l$, and $\odot$ denotes the Hadamard product.

Then, the node and edge embeddings are updated as follows:
\begin{align}
    h_i^{l+1} &= h_i^l + \text{ReLU}\left(\text{Norm}\left(U^l h_i^l + m_i^l\right)\right) \label{eq:node_update} \\
    \hat{e}_{ij}^{l+1} &= \hat{e}_{ij}^l + \text{ReLU}\left(\text{Norm}\left(A^l \hat{e}_{ij}^l + B^l h_i^l + C^l h_j^l\right)\right) \label{eq:edge_update}
\end{align}
where $U^l, A^l, B^l, C^l \in \mathbb{R}^{d \times d}$ are also learnable weight matrices of layer $l$, and $\text{Norm}$ represents Layer Normalization \cite{layernorm}.

To guarantee scale extrapolation, critical design decisions were made. First, the aggregation is performed via the $\text{MAX}$ operation, which is intrinsically agnostic to the number of nodes, preventing the embedding magnitudes from exploding in dense graphs. Second, Layer Normalization guarantees that the feature vectors remain in a predictable numerical range outside the training domain. After $L_\text{enc}$ convolution layers, the encoder outputs the final structural embeddings for all $N$ nodes in the graph, denoted as $H^{(L_\text{enc})} = \{h_i^{(L_\text{enc})}\}_{i=1}^N$, serving as the fundamental input for the autoregressive decoding mechanism.

\subsubsection{Autoregressive Attention Decoder} \label{subsubsec:decoder}
For route generation, we utilize an autoregressive decoder based on attention mechanisms \cite{kool2018attention}. This models the solution construction as a sequential process, capturing the permutation structure of the TSP.

At each time step $t$, the decoder generates a context vector $\hat{h}_t^C$ that integrates three sources of information: the global representation of the graph, the currently visited node, and the first selected node. This integration is defined by a linear projection of their concatenated embeddings:

\begin{equation}
    \hat{h}_t^C = W_C [h_G, h_{a_{t-1}}, h_{a_1}],
\end{equation}

where $h_G \in \mathbb{R}^d$ is the global graph embedding, derived by averaging the final node embeddings output by the encoder $H^{(L_\text{enc})}$ (i.e., $h_G = \frac{1}{N} \sum_{i=1}^N h_i^{(L_\text{enc})}$); $h_{a_{t-1}} \in \mathbb{R}^d$ represents the embedding of the currently visited node; and $h_{a_1} \in \mathbb{R}^d$ is the starting node of the route. The learnable projection matrix $W_C \in \mathbb{R}^{d \times 3d}$ maps the concatenated input to a embedding space, yielding the context vector $\hat{h}_t^C \in \mathbb{R}^d$. In the initial step ($t=1$), there is not route history, then, two trainable parameter vectors $v^{\text{last}}, v^{\text{first}} \in \mathbb{R}^d$ are used as placeholders.

This context vector $\hat{h}_t^C$ is processed by a Multi-Head Cross Attention (MHA) mechanism to evaluate the relevance of unvisited nodes. Following the standard attention formulation, we first obtain the query vector $q_{tm}$, the keys matrix $K_m$, and the values matrix $V_m$ for each attention head $m \in \{1, \dots, M\}$ by applying learnable linear projections to the context vector and the global node embeddings $H^{(L_\text{enc})}$:

\begin{align}
    q_{tm} &= (\hat{h}_t^C)^\top W_m^Q \\
    K_m &= H^{(L_\text{enc})} W_m^K \\
    V_m &= H^{(L_\text{enc})} W_m^V,
\end{align}

where $W_m^Q, W_m^K, W_m^V \in \mathbb{R}^{d \times d_k}$ are the projection weight matrices, and $d_k = d/M$ is the dimensionality of each attention head. With these components explicitly defined, the output of the attention head is computed using the standard scaled dot-product formulation:

\begin{equation}
    \text{head}_m = \text{softmax}\left( \frac{q_{tm} K_m^\top}{\sqrt{d_k}} \right) V_m.
\end{equation}

The outputs from all heads are concatenated and linearly projected to obtain the final refined context $\tilde{h}_t^C$:

\begin{equation}
    \tilde{h}_t^C = \text{Concat}(\text{head}_1, \dots, \text{head}_M) W^O,
\end{equation}

where $W^O \in \mathbb{R}^{d \times d}$. Finally, the selection probabilities (or logits), $u_{t}^{j}$, for the next node $j$ are obtained through a final single-head attention mechanism (acting as a pointer) that uses a hyperbolic tangent function as a non-linear function:

\begin{equation}
    u_{t}^{j} = \begin{cases} 
    C \cdot \tanh \left( \frac{Q_t^{\top} K_j}{\sqrt{d}} \right) & \text{if } j \notin \{ a_{1}, \dots, a_{t-1} \} \\ 
    -\infty & \text{otherwise} 
    \end{cases}
    \label{eq:pointer_attention}
\end{equation}
where the queries ($Q_t$) and keys ($K_j$) for this single-head attention are defined respectively as:
\begin{equation*}
    Q_t = W^Q_{\text{ptr}}\tilde{h}_{t}^{C} \quad \text{and} \quad K_j = W^K_{\text{ptr}}h_{j}^{(L_\text{enc})}.
\end{equation*}

Here, the dot product measures the similarity between the refined context $\tilde{h}_t^C$ and the structural node embedding $h_j^{(L_\text{enc})}$ provided by the encoder. The parameters $W^Q_{\text{ptr}}, W^K_{\text{ptr}} \in \mathbb{R}^{d \times d}$ are independent learnable projection matrices specific to this final pointer mechanism. The parameter $C$ (set to 10) clips the values to encourage exploration during training, and the value $-\infty$ (acting as a safety mask) strictly blocks previously visited nodes to guarantee a valid Hamiltonian permutation. 

The stochastic policy $\pi_{\theta}$ is obtained from these unnormalized compatibilities $u_{t}^{j}$ through a softmax function, yielding a valid probability distribution over the available nodes:

\begin{equation}
    \pi_{\theta}(a_t = j \mid \mathcal{G}, a_{<t}) = \frac{\exp(u_t^j)}{\sum_{k=1}^N \exp(u_t^k)}
\end{equation}

Finally, the action $a_t$ (i.e., the next node to visit) is extracted from this policy, and the specific procedure depends on the execution phase. During training, the next node is sampled from the distribution ($a_t \sim \pi_{\theta}$) to promote exploration of the state space. Conversely, during inference, a greedy decoding strategy is employed, deterministically selecting the node with the highest probability:

\begin{equation}
    a_t = \arg\max_j \pi_{\theta}(a_t = j \mid \mathcal{G}, a_{<t})
    \label{eq:3b}
\end{equation}

\subsection{Geometric Self-Supervised Pre-training}
\label{subsec:pretraining}

The fundamental limitation of training DRL routing agents from scratch is their susceptibility to geometric over-squashing and poor zero-shot extrapolation. Without a preliminary understanding of the topological space, the encoder tends to memorize the absolute coordinate distribution of the specific training scale rather than the underlying relative distance metrics. To mitigate this, we propose a self-supervised pre-training framework that forces the GatedGCN encoder to capture scale-invariant spatial symmetries prior to the policy optimization phase.

\begin{figure}[!t]
\centering
\subfloat[Original graph]{%
    \includegraphics[width=0.24\columnwidth]{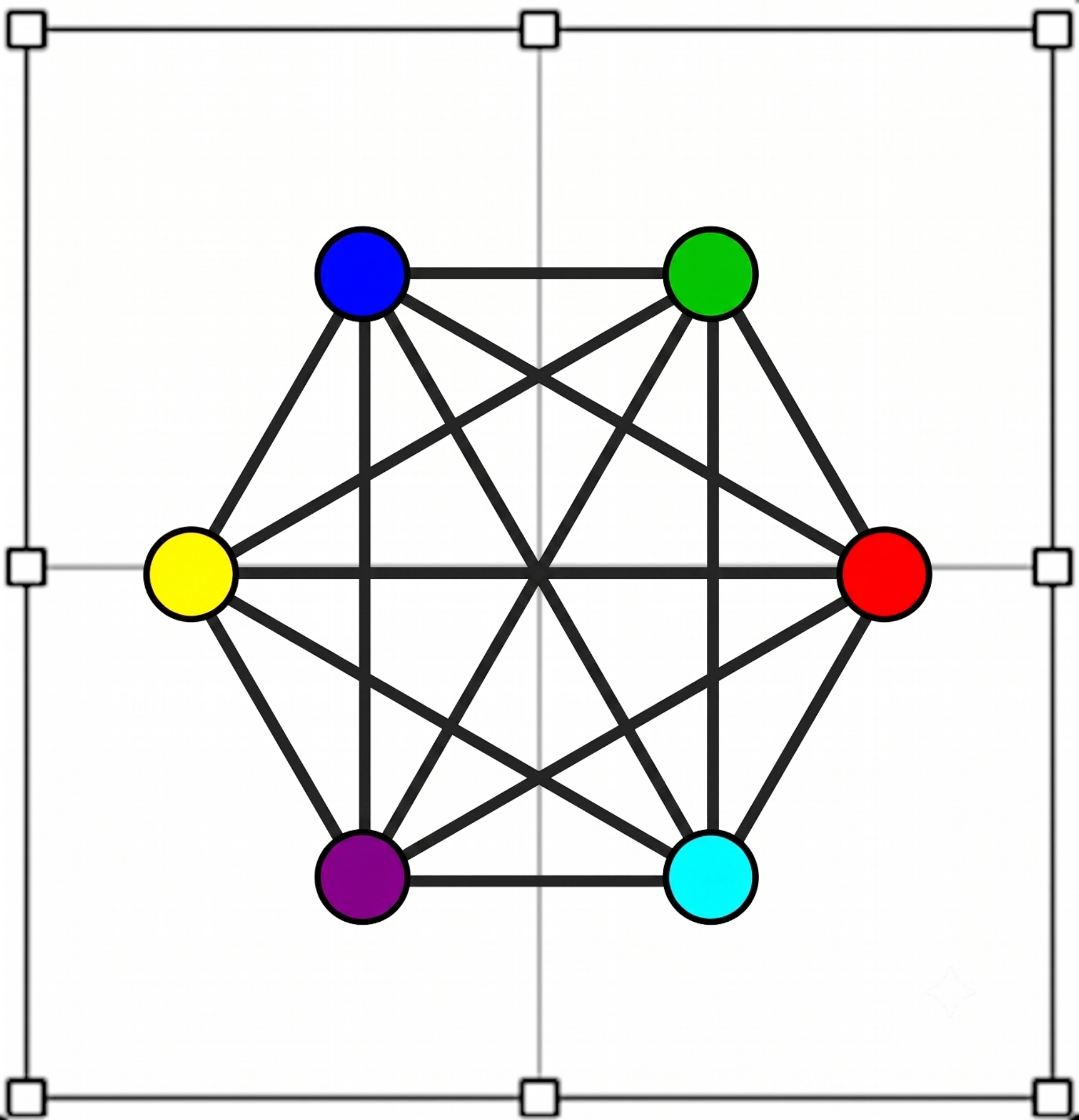}%
    \label{fig:trans_a}%
}
\hfil
\subfloat[Rotated graph]{%
    \includegraphics[width=0.245\columnwidth]{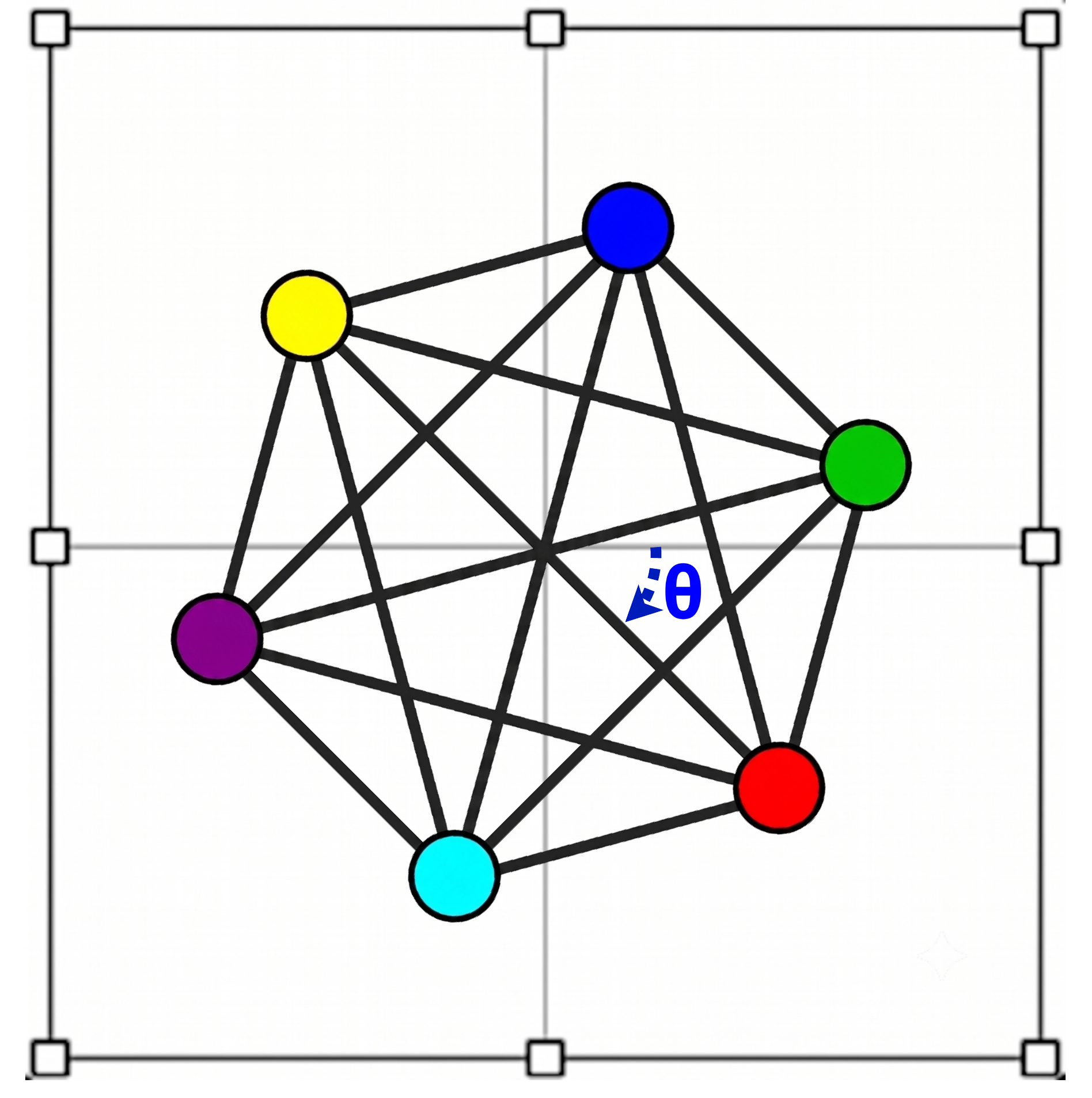}%
    \label{fig:trans_b}%
}

\vspace{1ex}

\subfloat[Reflected graph]{%
    \includegraphics[width=0.245\columnwidth]{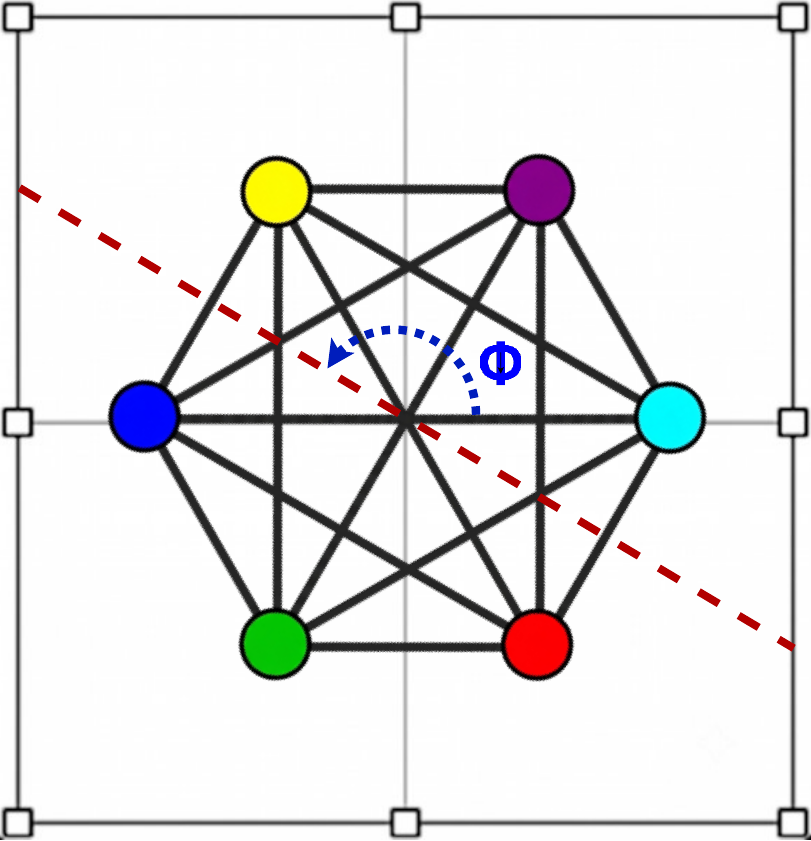}%
    \label{fig:trans_c}%
}
\hfil
\subfloat[Translated graph]{%
    \includegraphics[width=0.245\columnwidth]{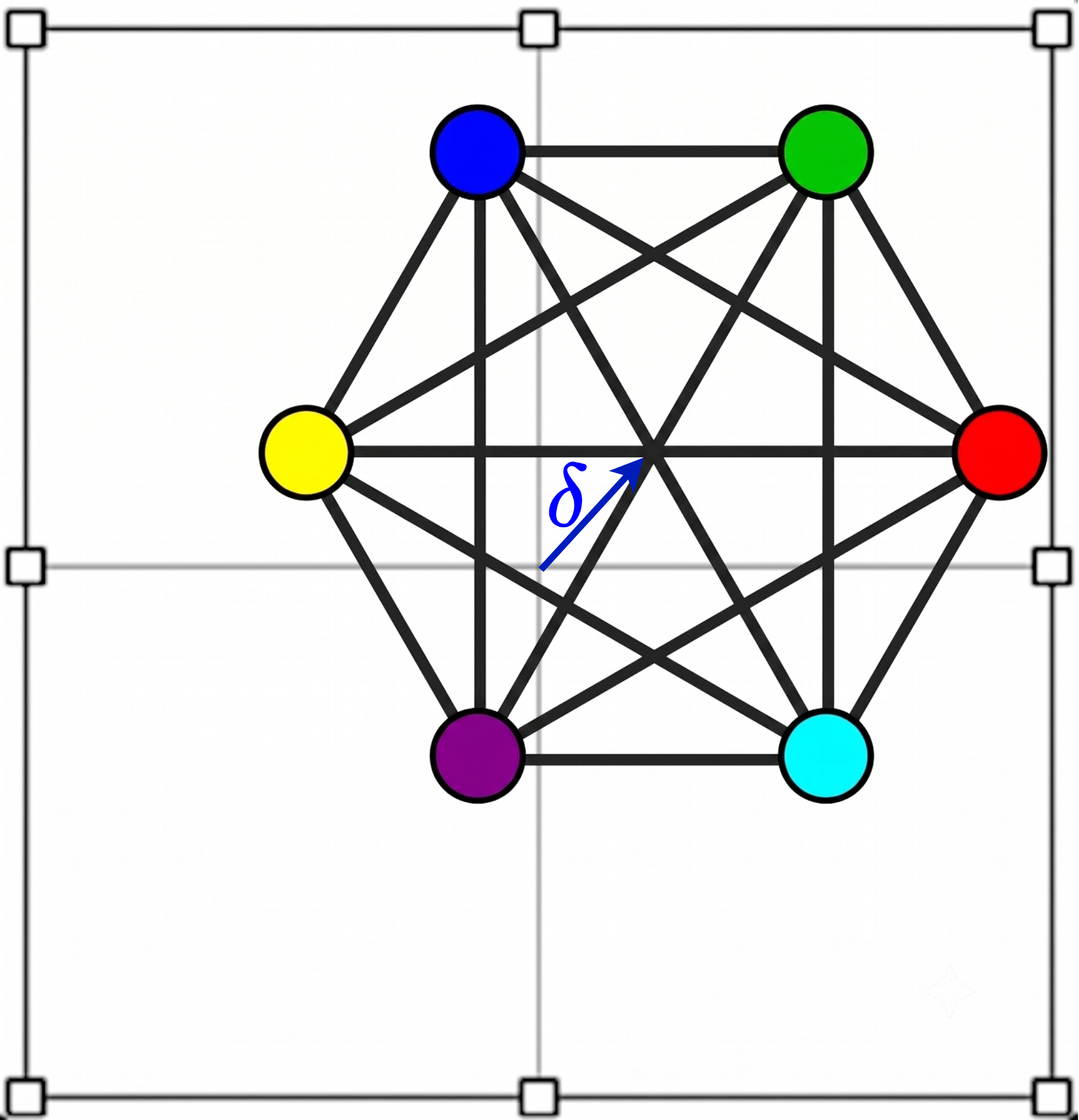}%
    \label{fig:trans_d}%
}
\caption{Geometric transformations for graph pre-training: (a) original instance, (b) rotation $f_\theta$, (c) axial reflection $f_\phi$, and (d) translation $f_\delta$.}
\label{fig:transformations_summary}
\end{figure}

Our approach is founded on a core geometric property of the Euclidean TSP: the optimal routing sequence remains strictly invariant under rigid isometric transformations that preserve the relative distances between all pairs of nodes. By leveraging contrastive learning \cite{contrastivelearning}, we train the encoder to maximize the agreement between different augmented views of the same underlying graph topology, a process conceptually depicted in the pre-training module of Figure \ref{fig:architecture}, using the spatial transformations detailed in Figure \ref{fig:transformations_summary}.

The choice of geometric augmentation transformations is critical for enabling the encoder to capture relative distance metrics $d_{ij}$ while ignoring absolute spatial coordinates. We implement three isometric transformations, represented in Figure \ref{fig:transformations_summary}, that operate on the node coordinates $x_i \in [0, 1]^2$ for a set of nodes $\mathcal{N} = \{1, \dots, n\}$: Central Rotation, Axial Reflection and Spatial Translation. In the Central Rotation, the graph is rotated around the geometric center $c = [0.5, 0.5]^\top$ of the unit square by an angle $\theta$ sampled from $\mathcal{U}[0, 2\pi)$. The transformed coordinates are:
\begin{equation}
    f_{\theta}(x_i) = R_{\theta}(x_i - c) + c, \quad R_{\theta} = \begin{bmatrix} \cos\theta & -\sin\theta \\ \sin\theta & \cos\theta \end{bmatrix}.
\end{equation}
In the Axial Reflection, the transformation mirrors the graph across an axis passing through $c$ with an orientation $\phi \sim \mathcal{U}[0, \pi)$, utilizing the Householder reflection matrix $S_{\phi}$ \cite{Householder}:
\begin{equation}
    f_{\phi}(x_i) = S_{\phi}(x_i - c) + c, \quad S_{\phi} = 2vv^\top - I
\end{equation}
where $v = [\cos\phi, \sin\phi]^\top$ is the unit vector defining the reflection axis, and $I$ is the identity matrix. Finally, in the Spatial Translation, a translation vector $\delta = [\delta_x, \delta_y]^\top$ is applied uniformly to all nodes:
\begin{equation}
    f_{\delta}(x_i) = x_i + \delta.
\end{equation}
To maintain numerical stability, the displacement $\delta$ is constrained to a maximum of 15\% of the total coordinate range, reinforcing the policy's independence from absolute coordinate origins.

Additionally, to maximize diversity in the graph contrastive views, we implement a stochastic Hybrid Transformation Strategy, consisting in the composition of previous transformations as
\begin{equation}
    g(x_i) = f_{\delta}^{b_{\delta}}(f_{\theta}^{b_{\theta}}(f_{\phi}^{b_{\phi}}(x_i))),
\end{equation}
where $b_{\phi}, b_{\theta}, b_{\delta} \sim \text{Bernoulli}(p)$ with $p=0.5$ are independent binary random variables. The conditional operator $f_{k}^{b_{k}}$ applies the transformation if $b_k=1$ and returns the identity otherwise:
\begin{equation}
    f_{k}^{b_{k}}(x) = \begin{cases} f_k(x) & \text{if } b_k = 1 \\ x & \text{if } b_k = 0 \end{cases}
\end{equation}
By applying these isometric transformations, we generate augmented views of the graph with significantly different coordinate representations but identical pairwise distances. This guarantees that the optimal TSP solution remains completely unchanged across views, forcing the encoder to extract deep, robust features that are invariant to spatial perturbations while preserving the underlying metric structure.

Once the transformations to be applied are defined, a cost or loss function is established to perform graph-level contrastive learning: global representations of the original graph and its augmented view should be compared. However, following standard contrastive learning frameworks, these global embeddings are not contrasted directly in the encoder's output space. Instead, an auxiliary projection head, consisting of a two-layer Multi-Layer Perceptron (MLP), temporarily maps the global graph embeddings into a lower-dimensional latent space. For the $i$-th instance in a batch, let $h_{G,i}$ and $h'_{G,i}$ be the global embeddings of the original and augmented views, respectively. Their latent projections are defined as $z_i = \text{MLP}(h_{G,i})$ and $z'_i = \text{MLP}(h'_{G,i})$.

To maximize the agreement between views of the same instance and minimize it against other instances, we apply the InfoNCE (Information Noise Contrastive Estimation) loss \cite{infoNCE}, which is formulated for a given batch of size $N_b$ as:

\begin{equation} 
\mathcal{L}_{\text{InfoNCE}} = - \frac{1}{N_b} \sum _{i=1}^{N_b} \log \frac{\exp(z_i^\top z'_i)}{\sum _{j=1}^{N_b} \exp(z_i^\top z'_j)},
\end{equation}

where the numerator represents the similarity of the positive pair $(z_i, z'_i)$, and the denominator sums the similarities between the anchor $z_i$ and all augmented views in the batch, acting as negative samples when $j \neq i$.

This objective function forces the encoder to capture connectivity patterns and distance distributions that are strictly essential for routing. After the pre-training is complete, the MLP projection head is discarded, allowing the RL phase to utilize the rich structural features retained in the unprojected embeddings ($h_G$).

\subsection{Reinforcement Learning Training}
\label{subsec:rl_optimization}

Following the contrastive pre-training phase, the model undergoes its primary training stage, as depicted in the overall architecture in Fig.\ref{fig:architecture}. Given that the TSP is an NP-hard combinatorial task, generating ground-truth optimal labels for large-scale instances is computationally prohibitive. Therefore, the system is trained end-to-end using DRL, allowing the model to learn routing policies directly from the reward signal defined by the final tour length.

We model the tour construction as a Markov Decision Process (MDP) \cite{MDP}, where the agent, represented by our neural architecture, sequentially selects nodes. The objective is to minimize the expected tour length $L(a)$ under the policy $\pi_\theta(a | \mathcal{G})$ using the loss function 
\begin{equation}
    \mathcal{L}(\theta | \mathcal{G}) = \mathbb{E}_{\pi_\theta(a | \mathcal{G})} [ L(a) ].
\end{equation}

To update the network parameters $\theta$, we employ the REINFORCE gradient estimator \cite{reinforce}. To mitigate the high variance inherent in the tour cost $L(a)$, we subtract a baseline $b(\mathcal{G})$ that approximates the expected cost of the current policy. The gradient of the loss is formulated as:
\begin{equation}
    \nabla_\theta \mathcal{L}(\theta | \mathcal{G}) = \mathbb{E}_{\pi_\theta(a | \mathcal{G})} \left[ (L(a) - b(\mathcal{G})) \nabla_\theta \log \pi_\theta(a | \mathcal{G}) \right].
\end{equation}
The baseline $b(\mathcal{G})$ is implemented following the Greedy Rollout Baseline \cite{kool2018attention} as the cost of a tour generated by a deterministic version of the model, where actions are selected according to equation \ref{eq:3b}. 

The sampled policy receives a positive update only when it outperforms the greedy baseline. The baseline is periodically replaced with the training model whenever a paired t-test ($\alpha=0.05$) confirms a statistically significant reduction in average tour length. Finally, the estimated gradients are processed by the Adam optimizer, which updates the network parameters to guide the model toward increasingly efficient routing policies.

\section{Results and Discussion} \label{sec:results}
This section presents the empirical evaluation of the proposed system, aiming to quantify the impact of different geometric pre-training strategies on the model's ability to solve the TSP. First, we conduct an ablation study to isolate the effects of individual and combined isometric transformations. Subsequently, we analyze the impact of scale variability during training on the model's zero-shot extrapolation capabilities, and finally, we benchmark our best-performing neural architecture against the state-of-the-art exact solver Concorde and some classical heuristic methods.

\subsection{Experimental Setup} \label{sec:experiments}
To ensure the rigor and reproducibility of the results presented in this paper, this section details the hardware environment, hyperparameters, evaluation datasets, and the performance metrics used during the learning and validation phases.

All experiments were conducted on a system equipped with an NVIDIA GeForce RTX 4090 GPU following a learning lifecycle divided into two consecutive methodological phases. In the first one, Pre-training Phase (Contrastive Learning), the encoder is configured with a latent embedding dimension of $128$. The self-supervised learning process spans $100$ epochs, processing $128,000$ unique graph instances per epoch, which are processed in mini-batches of $512$ instances. The network parameters are optimized using the Adam optimizer with a constant learning rate of $1 \times 10^{-4}$. To endow the encoder with a versatile inductive bias against structural variability, the number of nodes in the training graphs is dynamically sampled from a uniform distribution between $20$ and $50$ nodes. And in the second one, Training Phase (RL), the policy optimization via the REINFORCE algorithm is extended for $50$ additional epochs, maintaining the latent embedding dimension at $128$. The epoch size remains constant at $128,000$ instances, organized into batches of $256$ instances. The gradients are processed using the Adam optimizer, configured with a base learning rate of $10^{-4}$. 

For the validation, the model's performance is evaluated on independent datasets, each comprising $1,000$ unique TSP instances randomly generated in the 2D Euclidean plane. The quantitative evaluation framework relies on two primary metrics. The first one, Average Cost, calculates the mean tour length ($\mu$) and its sample standard deviation ($s$) over the test sets. And the second one, Optimality Gap, uses the exact solver Concorde as the performance baseline to compute the deviation of the neural model's solution quality as
\begin{equation}
    \text{GAP}_\text{TSP}(\%) = \frac{\mu_\text{NCO} - \mu_\text{Concorde}}{\mu_\text{Concorde}} \times 100,
\end{equation}
where $\mu_\text{NCO}$ and $\mu_\text{Concorde}$ denote the average costs obtained by our neural solver and Concorde, respectively. 

\subsection{Ablation Study and State-of-the-Art Comparison}
To empirically determine the contribution of the proposed isometric transformations and contextualize our results within the current literature, we conduct a joint analysis of solution quality and computational efficiency. All neural models evaluated in this section were pre-trained on instances of variable size (20 to 50 nodes) and subsequently fine-tuned via RL exclusively on 50-node graphs. Their zero-shot extrapolation capabilities were then tested on scales ranging from TSP20 up to massive TSP1,000 instances.

\begin{table}[htpb]
\centering
\caption{Evolution of the Optimality Gap (\%) for different pre-training strategies compared to Concorde.}
\label{tab:ablation_gap}
\resizebox{\columnwidth}{!}{%
\begin{tabular}{lcccccc}
\hline
\textbf{Pre-training Strategy} & \textbf{TSP20} & \textbf{TSP50} & \textbf{TSP100} & \textbf{TSP200} & \textbf{TSP500} & \textbf{TSP1000} \\ \hline
None (Baseline) & 4.6 & 6.9 & 13.8 & 25.1 & 46.0 & 67.2 \\
Rotation (Rot) & 3.7 & 6.0 & 12.2 & 22.8 & 40.7 & 58.2 \\
Reflection (Refl) & 3.7 & 5.9 & 12.0 & 21.4 & 40.3 & 57.5 \\
Translation (Trans) & 5.1 & 6.2 & 13.0 & 26.3 & 51.0 & 76.2 \\
Rot + Refl (Proposed) & \textbf{3.7} & 5.9 & \textbf{11.6} & \textbf{21.5} & \textbf{38.4} & \textbf{55.1} \\
Rot + Trans & 4.2 & \textbf{5.4} & 11.7 & 25.3 & 61.9 & 106.9 \\ 
Trans + Refl & 4.2 & 6.6 & 13.1 & 23.8 & 46.4 & 66.8 \\
Rot + Refl + Trans & 3.6 & 5.8 & 11.6 & 21.7 & 41.2 & 60.6 \\ \hline
\end{tabular}%
}
\end{table}

First, we isolate the effects of the geometric augmentations. Table \ref{tab:ablation_gap} details the evolution of the optimality gap against the exact solver Concorde for different pre-training strategies. The numerical metrics strongly corroborate the premise of geometric self-supervised learning: pre-training consistently establishes a generalized superiority over random initialization. While the baseline model trained from scratch suffers a severe performance degradation when extrapolating to massive instances, isolated pre-training strategies such as rotation or reflection successfully mitigate this steep increase. However, spatial translation has a highly detrimental impact when scaling the problem's dimensionality, yielding the worst overall results at TSP1,000. Our hypothesis is that translation introduces a detrimental geometric bias: it conditions the model to assume that a coordinate $(x, y)$ is functionally equivalent to a slightly displaced one, such as $(x + \delta, y + \delta)$. While this assumption has a negligible impact in low-density configurations, it becomes critical in massive, highly dense graphs, where even a minimal spatial variation separates two completely distinct nodes. By erroneously interpreting these extremely close locations as analogous, the network loses its ability to discriminate strict routing precedence, which severely hinders its capacity to resolve local neighborhoods. Ultimately, the hybrid composition of rotation and reflection emerges as the dominant strategy, confirming that combining strictly distance-preserving transformations without spatial shifts is highly effective.

Second, we benchmark our best-performing hybrid model (Rotation + Reflection) against the baseline NCO without pre-training, the exact solver Concorde, and classical expert-designed heuristics. Table \ref{tab:sota_comparison} presents a comprehensive comparison of both the optimality gap and the computational inference time. 

This joint analysis highlights a fundamental operational trade-off. Concorde guarantees the optimal Hamiltonian cycle; however, due to the NP-hard nature of the TSP, its execution time grows exponentially, creating a severe computational bottleneck. Classical heuristics, such as LKH-3, offer high-quality solutions but still exhibit a noticeable increase in computational overhead at massive scales. In contrast, our pre-trained NCO framework achieves the best balance between solution quality and computational cost. While trading absolute optimality for speed in massive extrapolation scenarios, it maintains a near-instantaneous execution profile, delivering speedups of up to two orders of magnitude over Concorde at 1,000 nodes. 

\begin{table*}[htpb]
\centering
\caption{State-of-the-Art Comparison: Optimality Gap (\%) and Inference Time (s) across increasing TSP scales.}
\label{tab:sota_comparison}
\resizebox{\textwidth}{!}{%
\begin{tabular}{l|cc|cc|cc|cc|cc|cc}
\hline
\multirow{2}{*}{\textbf{Algorithm}} & \multicolumn{2}{c|}{\textbf{TSP20}} & \multicolumn{2}{c|}{\textbf{TSP50}} & \multicolumn{2}{c|}{\textbf{TSP100}} & \multicolumn{2}{c|}{\textbf{TSP200}} & \multicolumn{2}{c|}{\textbf{TSP500}} & \multicolumn{2}{c}{\textbf{TSP1000}} \\
 & Gap (\%) & Time (s) & Gap (\%) & Time (s) & Gap (\%) & Time (s) & Gap (\%) & Time (s) & Gap (\%) & Time (s) & Gap (\%) & Time (s) \\ \hline
Concorde (Exact) & 0.0 & 0.06341 & 0.0 & 0.11 & 0.0 & 0.35 & 0.0 & 1.42 & 0.0 & 14.39 & 0.0 & 175.19 \\
LKH-3 & 0.0 & 0.05412 & 0.0 & 0.06360 & 0.0 & 0.09787 & 0.0 & 0.27 & 0.0 & 1.61 & 0.0 & 7.18 \\
Google OR-Tools & 0.3 & 0.05169 & 2.7 & 0.05160 & 4.5 & 0.10 & 5.3 & 0.50 & 4.4 & 7.01 & 5.1 & 80.06 \\ 
Ant Colony Optimization & 1.5 & 2.69 & 2.4 & 26.65 & 3.9 & 181.56 & - & - & - & - & - & - \\
Genetic Algorithm & 1.7 & 1.08 & 0.8 & 14.41 & 3.4 & 144.56 & - & - & - & - & - & - \\ \hline
NCO Baseline (No Pre-train) & 4.6 & 0.00026 & 6.9 & 0.00077 & 13.8 & 0.00353 & 25.1 & 0.01260 & 46.0 & 0.09417 & 67.2 & 0.499 \\
\textbf{NCO Proposed (Rot+Refl)} & \textbf{3.7} & \textbf{0.00029} & \textbf{5.9} & \textbf{0.00082} & \textbf{11.6} & \textbf{0.00361} & \textbf{21.5} & \textbf{0.01269} & \textbf{38.4} & \textbf{0.09428} & \textbf{55.1} & \textbf{0.50} \\ \hline
\end{tabular}%
}
\end{table*}

\begin{table*}[!t]
    \caption{Impact of training and pre-training with scale variability on zero-shot generalization and average tour costs.}
    \centering
    \resizebox{\textwidth}{!}{
    \begin{tabular}{|l|l|c|c|c|c|c|c|}
        \hline
        \multirow{2}{*}{\textbf{Pre-training}} & \multirow{2}{*}{\textbf{Training}} & \multicolumn{6}{c|}{\textbf{Evaluation}} \\
        \cline{3-8}
        & & \textbf{TSP 20} & \textbf{TSP 50} & \textbf{TSP 100} & \textbf{TSP 200} & \textbf{TSP 500} & \textbf{TSP 1,000} \\
        \hline
        20-50 & 20-50 & $\mathbf{3.922 \pm 0.020}$ & $6.071 \pm 0.019$ & $8.846 \pm 0.033$ & $13.561 \pm 0.071$ & $25.852 \pm 0.088$ & $43.420 \pm 0.137$ \\
        \hline
        20-50 & 50 & $3.949 \pm 0.020$ & $6.028 \pm 0.019$ & $8.626 \pm 0.033$ & $12.959 \pm 0.065$ & $23.350 \pm 0.089$ & $37.405 \pm 0.137$ \\
        \hline
        20-100 & 20-100 & $\mathbf{3.909 \pm 0.020}$ & $\mathbf{5.963 \pm 0.019}$ & $\mathbf{8.425 \pm 0.033}$ & $\mathbf{12.334 \pm 0.075}$ & $21.686 \pm 0.087$ & $34.192 \pm 0.147$ \\
        \hline
        50 & 50 & $3.957 \pm 0.021$ & $\mathbf{5.990 \pm 0.019}$ & $8.545 \pm 0.033$ & $12.927 \pm 0.067$ & $23.927 \pm 0.091$ & $38.635 \pm 0.133$ \\
        \hline
        100 & 100 & $4.578 \pm 0.038$ & $6.312 \pm 0.019$ & $8.558 \pm 0.033$ & $\mathbf{12.225 \pm 0.082}$ & $\mathbf{20.931 \pm 0.087}$ & $\mathbf{32.553 \pm 0.139}$ \\
        \hline
    \end{tabular}
    }
    \label{tab:scale_variability}
\end{table*}

To visually illustrate the qualitative impact of our geometric approach, Appendix \ref{sec:apendice} provides a side-by-side comparison of the routing sequences generated by the baseline model, various pre-training configurations, and the optimal solver Concorde. This visual analysis demonstrates the structural coherence gained through the different pre-training strategies across varying instance sizes.

\subsection{Scalability and Zero-Shot Extrapolation}
\label{subsec:scalability}
A fundamental challenge in NCO is the tendency of DRL agents to overfit to the specific graph size encountered during training. When these models are evaluated on significantly larger instances without prior fine-tuning, their performance typically degrades drastically. This collapse occurs because the network internalizes the specific node density and distance distributions of the training set, failing to generalize its routing policy when exposed to the denser topologies of massive graphs. To counteract this scale-dependent bias, our methodology introduces scale variability during the learning phases. Rather than exposing the network to a static graph size (e.g., exactly 50 nodes), the model is trained on instances dynamically sampled from a uniform distribution of sizes (e.g., $N \sim \mathcal{U}(20, 50)$). This stochastic variation in the number of nodes forces the GatedGCN encoder and the attention mechanisms to learn scale-agnostic representations. Consequently, the model must rely on the underlying geometric relationships and relative distances rather than memorizing the absolute topological characteristics of a fixed-size environment.

The quantitative impact of introducing this dynamic sizing is detailed in Table \ref{tab:scale_variability}. The results demonstrate a clear correlation between scale variability during the pre-training/training phases and the model's resilience in zero-shot extrapolation scenarios. When models are restricted to a strictly fixed training scale, they suffer a steep increase in the average tour costs when confronted with massive instances such as TSP500 or TSP1,000. The encoder, accustomed to a specific spatial density, becomes disoriented by the sheer number of neighboring nodes in high-dimensional spaces.

Conversely, incorporating scale variability acts as a powerful structural regularizer. When this variable-scale approach is combined with our optimal hybrid geometric pre-training (Rotation + Reflection), the degradation of the routing policy is significantly contained up to moderate extrapolation scales. 

However, an important nuance emerges from the empirical data regarding extreme extrapolation. While scale flexibility proves highly beneficial for transitioning between moderate sizes, the results indicate that for scenarios of ultra-high node density, the benefit of scale flexibility dilutes in favor of structural specialization. In massive, highly dense instances (such as TSP1,000), the topological complexity and the sheer number of local neighborhoods demand a highly specialized feature extraction. In these extreme environments, a model trained on a fixed, specific scale may capitalize on its ``over-specialization'' to resolve localized clusters better than a generalized, scale-agnostic model. This reveals a critical trade-off in NCO design between maintaining universal scale flexibility and achieving peak performance through structural specialization in ultra-dense domains.

\section{Conclusions} \label{sec:conclusions}
In this paper, we address the generalization bottleneck of NCO in routing problems by introducing a novel geometric self-supervised pre-training framework. Our findings demonstrate that applying a carefully selected composition of isometric transformations forces the encoder to learn robust, scale-invariant spatial representations, effectively mitigating the structural collapse typically observed in baseline models during zero-shot extrapolation. Furthermore, while incorporating scale variability during training acts as a powerful regularizer for generalizing across moderate sizes, we identify a critical trade-off where ultra-dense instances ultimately require fixed-scale structural specialization to achieve peak performance. The proposed neural solver maintains near-instantaneous inference times, delivering speedups of up to two orders of magnitude over the exact solver Concorde and establishing a highly scalable alternative for real-time, industrial routing applications.

\section*{Acknowledgments}
This work was supported in part by the Comunidad de Madrid under project TEC-2024/COM-322 (IDEALCVCM), in part by MCIU/AEI/10.13039/501100011033 of the Spanish Government under project PID2023148922OA-I00 (EEVOCATIONS), and in part by ``Ayudas a la Investigación para el Personal Docente e Investigador de la ETSIT-UPM (2026)'' under project ``SATURNO''. The authors would also like to thank Airbus Defence and Space for their support.

\bibliographystyle{unsrt}  
\bibliography{references}  

@inproceedings{
kool2018attention,
title={Attention, Learn to Solve Routing Problems!},
author={Wouter Kool and Herke van Hoof and Max Welling},
booktitle={International Conference on Learning Representations},
year={2019},
}

@misc{GCN,
      title={Semi-Supervised Classification with Graph Convolutional Networks}, 
      author={Thomas N. Kipf and Max Welling},
      year={2017},
      eprint={1609.02907},
      archivePrefix={arXiv},
      primaryClass={cs.LG},
      url={https://arxiv.org/abs/1609.02907}, 
}

@misc{GAT,
      title={Graph Attention Networks}, 
      author={Petar Veličković and Guillem Cucurull and Arantxa Casanova and Adriana Romero and Pietro Liò and Yoshua Bengio},
      year={2018},
      eprint={1710.10903},
      archivePrefix={arXiv},
      primaryClass={stat.ML},
      url={https://arxiv.org/abs/1710.10903}, 
}

@INPROCEEDINGS{mae,
  author={He, Kaiming and Chen, Xinlei and Xie, Saining and Li, Yanghao and Dollár, Piotr and Girshick, Ross},
  booktitle={2022 IEEE/CVF Conference on Computer Vision and Pattern Recognition (CVPR)}, 
  title={Masked Autoencoders Are Scalable Vision Learners}, 
  year={2022},
  volume={},
  number={},
  pages={15979-15988},
  doi={10.1109/CVPR52688.2022.01553}}

@inproceedings{contrastivelearning,
author = {Chen, Ting and Kornblith, Simon and Norouzi, Mohammad and Hinton, Geoffrey},
title = {A simple framework for contrastive learning of visual representations},
year = {2020},
publisher = {JMLR.org},
booktitle = {Proceedings of the 37th International Conference on Machine Learning},
articleno = {149},
numpages = {11},
series = {ICML'20}
}

@BOOK{concorde,
 ISBN = {9780691129938},
 URL = {http://www.jstor.org/stable/j.ctt7s8xg},
 author = {David L. Applegate and Robert E. Bixby and Vašek Chvatál and William J. Cook},
 publisher = {Princeton University Press},
 title = {The Traveling Salesman Problem: A Computational Study},
 urldate = {2026-02-09},
 year = {2006}
}

@inproceedings{LKH-3,
  title={An Extension of the Lin-Kernighan-Helsgaun TSP Solver for Constrained Traveling Salesman and Vehicle Routing Problems: Technical report},
  author={Keld Helsgaun},
  year={2017},
  url={https://api.semanticscholar.org/CorpusID:57634432}
}

@article{ExplainTSP,
author = {Miller, C. E. and Tucker, A. W. and Zemlin, R. A.},
title = {Integer Programming Formulation of Traveling Salesman Problems},
year = {1960},
issue_date = {Oct. 1960},
publisher = {Association for Computing Machinery},
address = {New York, NY, USA},
volume = {7},
number = {4},
issn = {0004-5411},
url = {https://doi.org/10.1145/321043.321046},
doi = {10.1145/321043.321046},
journal = {J. ACM},
pages = {326–329},
numpages = {4}
}

@misc{GatedGCN,
      title={Gated Graph Sequence Neural Networks}, 
      author={Yujia Li and Daniel Tarlow and Marc Brockschmidt and Richard Zemel},
      year={2017},
      eprint={1511.05493},
      archivePrefix={arXiv},
      primaryClass={cs.LG},
      url={https://arxiv.org/abs/1511.05493}, 
}

@article{MDP,
title = {Markov decision processes},
journal = {European Journal of Operational Research},
volume = {39},
number = {1},
pages = {1-16},
year = {1989},
issn = {0377-2217},
doi = {https://doi.org/10.1016/0377-2217(89)90348-2},
url = {https://www.sciencedirect.com/science/article/pii/0377221789903482},
author = {Chelsea C. White and Douglas J. White},
}

@article{REINFORCE,
  author  = {Williams, Ronald J.},
  title   = {Simple statistical gradient-following algorithms for connectionist reinforcement learning},
  journal = {Machine Learning},
  year    = {1992},
  month   = {May},
  volume  = {8},
  number  = {3},
  pages   = {229--256},
  doi     = {10.1007/BF00992696},
  url     = {https://doi.org/10.1007/BF00992696},
  issn    = {1573-0565}
}

@article{2op,
author = {Astoquillca-Yaranga, David and Berger-Vidal, Esther},
year = {2023},
month = {06},
pages = {65-81},
title = {Heurística de Intercambio 2Opt Best Improvement y Nivel de Eficacia de las soluciones del Problema del Agente Viajero Simétrico},
volume = {5},
journal = {Revista peruana de computación y sistemas},
doi = {10.15381/rpcs.v5i1.25806}
}

@inproceedings{bert,
title	= {BERT: Pre-training of Deep Bidirectional Transformers for Language Understanding},
author	= {Jacob Devlin and Ming-Wei Chang and Kenton Lee and Kristina N. Toutanova},
year	= {2018},
URL	= {https://arxiv.org/abs/1810.04805}
}

@inproceedings{pointer,
 author = {Vinyals, Oriol and Fortunato, Meire and Jaitly, Navdeep},
 booktitle = {Advances in Neural Information Processing Systems},
 editor = {C. Cortes and N. Lawrence and D. Lee and M. Sugiyama and R. Garnett},
 pages = {},
 publisher = {Curran Associates, Inc.},
 title = {Pointer Networks},
 url = {https://proceedings.neurips.cc/paper_files/paper/2015/file/29921001f2f04bd3baee84a12e98098f-Paper.pdf},
 volume = {28},
 year = {2015}
}

@inproceedings{mae2,
author = {Hou, Zhenyu and He, Yufei and Cen, Yukuo and Liu, Xiao and Dong, Yuxiao and Kharlamov, Evgeny and Tang, Jie},
title = {GraphMAE2: A Decoding-Enhanced Masked Self-Supervised Graph Learner},
year = {2023},
isbn = {9781450394161},
publisher = {Association for Computing Machinery},
address = {New York, NY, USA},
url = {https://doi.org/10.1145/3543507.3583379},
doi = {10.1145/3543507.3583379},
booktitle = {Proceedings of the ACM Web Conference 2023},
pages = {737–746},
numpages = {10},
location = {Austin, TX, USA},
series = {WWW '23}
}

@inproceedings{maee,
author = {Hou, Zhenyu and Liu, Xiao and Cen, Yukuo and Dong, Yuxiao and Yang, Hongxia and Wang, Chunjie and Tang, Jie},
title = {GraphMAE: Self-Supervised Masked Graph Autoencoders},
year = {2022},
isbn = {9781450393850},
publisher = {Association for Computing Machinery},
address = {New York, NY, USA},
url = {https://doi.org/10.1145/3534678.3539321},
doi = {10.1145/3534678.3539321},
booktitle = {Proceedings of the 28th ACM SIGKDD Conference on Knowledge Discovery and Data Mining},
pages = {594–604},
numpages = {11},
location = {Washington DC, USA},
series = {KDD '22}
}

@article{branchandbound,
title = {Branch-and-bound algorithms: A survey of recent advances in searching, branching, and pruning},
journal = {Discrete Optimization},
volume = {19},
pages = {79-102},
year = {2016},
issn = {1572-5286},
doi = {https://doi.org/10.1016/j.disopt.2016.01.005},
url = {https://www.sciencedirect.com/science/article/pii/S1572528616000062},
author = {David R. Morrison and Sheldon H. Jacobson and Jason J. Sauppe and Edward C. Sewell},
}

@article{Householder,
author = {Householder, Alston S.},
title = {Unitary Triangularization of a Nonsymmetric Matrix},
year = {1958},
issue_date = {Oct. 1958},
publisher = {Association for Computing Machinery},
address = {New York, NY, USA},
volume = {5},
number = {4},
issn = {0004-5411},
url = {https://doi.org/10.1145/320941.320947},
doi = {10.1145/320941.320947},
journal = {J. ACM},
month = oct,
pages = {339–342},
numpages = {4}
}

@article{augmentation,
author = {Zhou, Jiajun and Xie, Chenxuan and Gong, Shengbo and Wen, Zhenyu and Zhao, Xiangyu and Xuan, Qi and Yang, Xiaoniu},
title = {Data Augmentation on Graphs: A Technical Survey},
year = {2025},
issue_date = {November 2025},
publisher = {Association for Computing Machinery},
address = {New York, NY, USA},
volume = {57},
number = {11},
issn = {0360-0300},
url = {https://doi.org/10.1145/3732282},
doi = {10.1145/3732282},
journal = {ACM Comput. Surv.},
month = jun,
articleno = {274},
numpages = {34}
}

@article{NCOreview,
  author   = {Chung, K. T. and Lee, C. K. M. and Tsang, Y. P.},
  title    = {Neural combinatorial optimization with reinforcement learning in industrial engineering: a survey},
  journal  = {Artificial Intelligence Review},
  year     = {2025},
  volume   = {58},
  number   = {5},
  pages    = {130},
  doi      = {10.1007/s10462-024-11045-1},
  url      = {https://doi.org/10.1007/s10462-024-11045-1},
  issn     = {1573-7462},
}

@article{DRLNCO,
  title={Neural Combinatorial Optimization with Heavy Decoder: Toward Large Scale Generalization},
  author={Fu Luo and Xi Lin and Fei Liu and Qingfu Zhang and Zhenkun Wang},
  journal={ArXiv},
  year={2023},
  volume={abs/2310.07985},
  url={https://api.semanticscholar.org/CorpusID:263909317}
}

@inproceedings{infoNCE,
title={Contrastive Difference Predictive Coding},
author={Chongyi Zheng and Ruslan Salakhutdinov and Benjamin Eysenbach},
booktitle={The Twelfth International Conference on Learning Representations},
year={2024},
url={https://openreview.net/forum?id=0akLDTFR9x}
}

@inproceedings{RRNCO,
title={{RRNCO}: Towards Real-World Routing with Neural Combinatorial Optimization},
author={Jiwoo Son and Zhikai Zhao and Federico Berto and Chuanbo Hua and Zhiguang Cao and Changhyun Kwon and Jinkyoo Park},
booktitle={Workshop on Differentiable Learning of Combinatorial Algorithms},
year={2025},
url={https://openreview.net/forum?id=MGLt2k07KC}
}

@article{generalization,
author = {Joshi, Chaitanya K. and Cappart, Quentin and Rousseau, Louis-Martin and Laurent, Thomas},
title = {Learning the travelling salesperson problem requires rethinking generalization},
year = {2022},
issue_date = {Apr 2022},
publisher = {Kluwer Academic Publishers},
address = {USA},
volume = {27},
number = {1–2},
issn = {1383-7133},
url = {https://doi.org/10.1007/s10601-022-09327-y},
doi = {10.1007/s10601-022-09327-y},
journal = {Constraints},
month = apr,
pages = {70–98},
numpages = {29}
}

@inproceedings{dgi,
title={Deep Graph Infomax},
author={Petar Veličković and William Fedus and William L. Hamilton and Pietro Liò and Yoshua Bengio and R Devon Hjelm},
booktitle={International Conference on Learning Representations},
year={2019},
url={https://openreview.net/forum?id=rklz9iAcKQ},
}

@InProceedings{graphcl,
  title = 	 {Graph Contrastive Learning Automated},
  author =       {You, Yuning and Chen, Tianlong and Shen, Yang and Wang, Zhangyang},
  booktitle = 	 {Proceedings of the 38th International Conference on Machine Learning},
  pages = 	 {12121--12132},
  year = 	 {2021},
  editor = 	 {Meila, Marina and Zhang, Tong},
  volume = 	 {139},
  series = 	 {Proceedings of Machine Learning Research},
  month = 	 {18--24 Jul},
  publisher =    {PMLR},
  url = 	 {https://proceedings.mlr.press/v139/you21a.html},
}

@inproceedings{layernorm,
author = {Xiong, Ruibin and Yang, Yunchang and He, Di and Zheng, Kai and Zheng, Shuxin and Xing, Chen and Zhang, Huishuai and Lan, Yanyan and Wang, Liwei and Liu, Tie-Yan},
title = {On layer normalization in the transformer architecture},
year = {2020},
publisher = {JMLR.org},
booktitle = {Proceedings of the 37th International Conference on Machine Learning},
articleno = {975},
numpages = {10},
series = {ICML'20}
}

@article{gao2023topology,
  author  = {Gao, Xiang and Hu, Wei and Qi, Guo-Jun},
  title   = {Self-Supervised Graph Representation Learning via Topology Transformations},
  journal = {IEEE Transactions on Knowledge and Data Engineering},
  year    = {2023},
  volume  = {35},
  number  = {4},
  pages   = {4202--4215},
  month   = apr,
  doi     = {10.1109/TKDE.2021.3133439}
}

@article{zhang2024personalized,
  author  = {Zhang, Xin and Tan, Qiaoyu and Huang, Xiao and Li, Bo},
  title   = {Graph Contrastive Learning With Personalized Augmentation},
  journal = {IEEE Transactions on Knowledge and Data Engineering},
  year    = {2024},
  volume  = {36},
  number  = {11},
  pages   = {6305--6316},
  month   = nov,
  doi     = {10.1109/TKDE.2024.3388728}
}

@article{li2024aegcl,
  author  = {Li, Wen-Zhi and Wang, Chang-Dong and Lai, Jian-Huang and Yu, Philip S.},
  title   = {Towards Effective and Robust Graph Contrastive Learning With Graph Autoencoding},
  journal = {IEEE Transactions on Knowledge and Data Engineering},
  year    = {2024},
  volume  = {36},
  number  = {2},
  pages   = {868--881},
  month   = feb,
  doi     = {10.1109/TKDE.2023.3288280}
}

@article{zhang2024multiscale,
  author  = {Zhang, Haonan and Ren, Yuyang and Fu, Luoyi and Wang, Xinbing and Chen, Guihai and Zhou, Chenghu},
  title   = {Multi-Scale Self-Supervised Graph Contrastive Learning With Injective Node Augmentation},
  journal = {IEEE Transactions on Knowledge and Data Engineering},
  year    = {2024},
  volume  = {36},
  number  = {1},
  pages   = {261--274},
  month   = jan,
  doi     = {10.1109/TKDE.2023.3278463}
}

@article{zhang2025gigraph,
  author  = {Zhang, Sanfeng and Liu, Xinyi and Qi, Zihao and Yan, Xingchen and Yang, Wang},
  title   = {{GI-Graph}: A Generative Invariant Graph Learning Scheme Towards Out-of-Distribution Generalization},
  journal = {IEEE Transactions on Knowledge and Data Engineering},
  year    = {2025},
  volume  = {37},
  number  = {10},
  pages   = {5934--5947},
  month   = oct,
  doi     = {10.1109/TKDE.2025.3592640}
}

\newpage

\appendix
\section{Visual comparison}\label{sec:apendice}
\begin{figure}[h]
\centering
\includegraphics[scale=0.155]{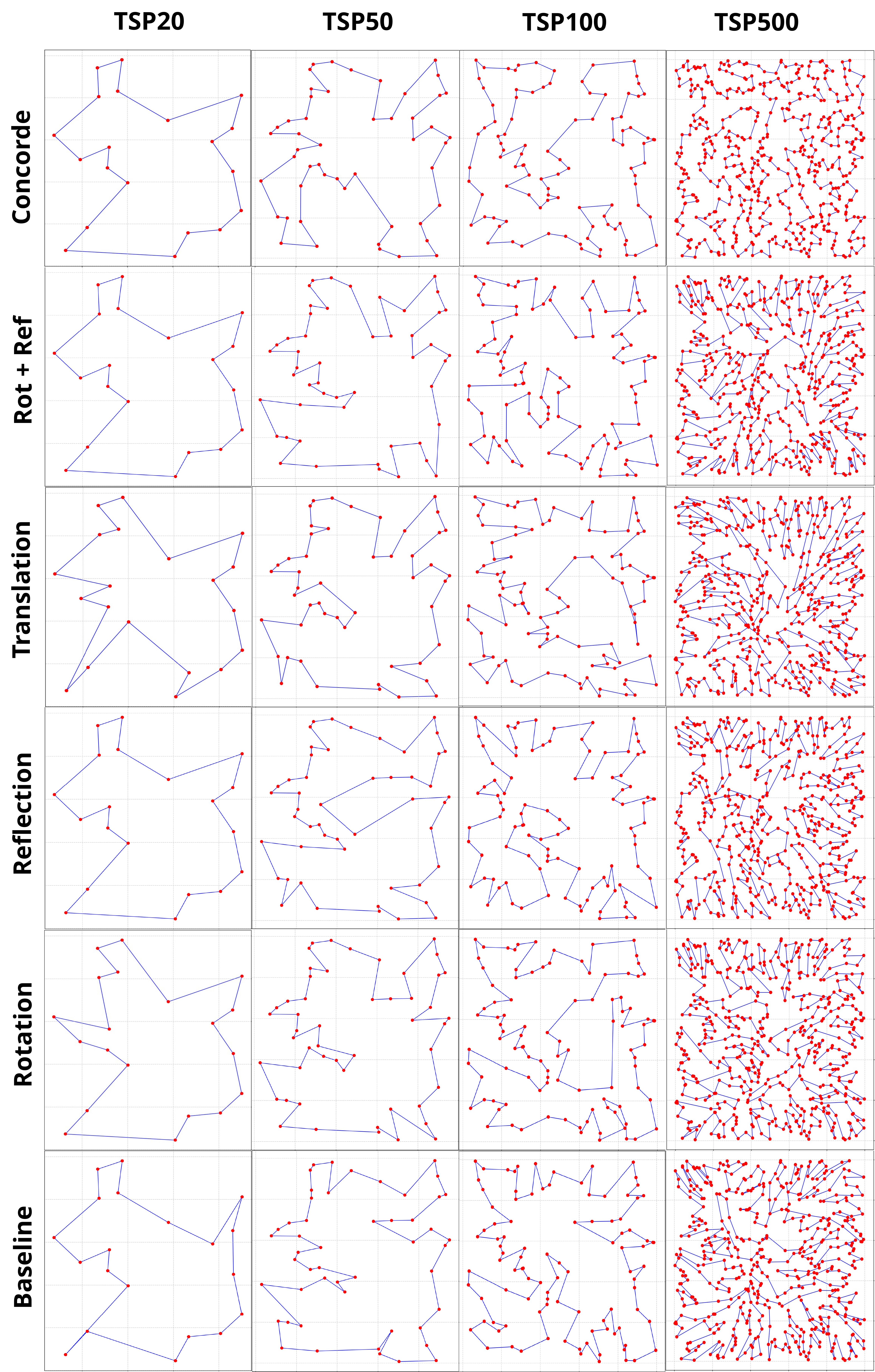}
\caption{Visual comparison of the routes generated by different architectures in response to progressive increases in the TSP scale.}
\label{fig:comp}
\end{figure}

\end{document}